\documentclass[runningheads]{llncs}
\usepackage[T1]{fontenc}
\usepackage{graphicx}
\usepackage{booktabs}
\usepackage[misc]{ifsym}

\usepackage{hyperref}
\usepackage{todonotes}
\makeatletter
\newcommand{\printfnsymbol}[1]{%
  \textsuperscript{\@fnsymbol{#1}}%
}
\makeatother

\begin{document}

\title{DetoxAI: a Python Toolkit for Debiasing Deep Learning Models in Computer Vision}

\titlerunning{ }
% If the full title of your paper is short enough to also fit in the running head, you can omit the abbreviated paper title here. You can check as follows: if you comment out the \titlerunning line, something will appear in the header of all odd-numbered pages of your PDF from page 3 onward. This something is either the full title (in which case all is well), or the error message "Title Suppressed Due to Excessive Length". If this error message appears, you're going to want to provide an abbreviated title within the \titlerunning command, because if you won't do it, Springer will do it for you.

\authorrunning{ }
%N.B.: Author information (both in the \author{} and \authorrunning{} command) should only be present in the Camera-Ready Version of your paper. The version that you initially submit for review, ought to be double-blind. So, when initially submitting your paper, use:
% \author{Author information scrubbed for double-blind reviewing}
\author{Ignacy St\k{e}pka \inst{1}\thanks{Equal contribution} \and
Lukasz Sztukiewicz \inst{1}\textsuperscript{$\star$}
 \and
Micha\l{} Wili\'{n}ski \inst{1}\textsuperscript{$\star$} \and 
Jerzy Stefanowski \inst{1} 
}

% You may leave out the orcidID information, if you want to.
% Use \corr to indicate the corresponding author. Note the spacing around the \corr command. Only one author can be the corresponding author.

% \institute{}
% \email{}
\institute{Institute of Computing Science, Poznan University of Technology}

%N.B.: comment out the \authorrunning{} command for the double-blind version of your paper submitted for review. Later, if your paper is accepted, use the command for the Camera-Ready Version.
% \authorrunning{St\k{e}pka et al.}
% If there are two authors, write 'A.L. Benjamin and C.C. Broadus Jr.'
% If there are more than two authors, '[...] et al.' is used.

\maketitle              % typeset the header of the contribution

\begin{abstract}
While machine learning fairness has made significant progress in recent years, most existing solutions focus on tabular data and are poorly suited for vision-based classification tasks, which rely heavily on deep learning. To bridge this gap, we introduce DetoxAI, an open-source Python library for improving fairness in deep learning vision classifiers through post-hoc debiasing. DetoxAI implements state-of-the-art debiasing algorithms, fairness metrics, and visualization tools. It supports debiasing via interventions in internal representations and includes attribution-based visualization tools and quantitative algorithmic fairness metrics to show how bias is mitigated. This paper presents the motivation, design, and use cases of DetoxAI, demonstrating its tangible value to engineers and researchers.

% \keywords{Fairness \and Deep Learning \and Computer Vision \and Debiasing}
\end{abstract}

\section{Introduction}

Ensuring fairness in machine learning models has become critical, particularly in high-stakes fields~\cite{mehrabi_2021_survey}. While several libraries address fairness, most focus on tabular data and are ill-suited for unstructured, high-dimensional tasks like computer vision. We identify two major gaps in the current landscape.

The first is technical: existing tools such as AIF360~\cite{aif360} and Fairlearn~\cite{fairlearn} are built around the scikit-learn API, expecting datasets to fit in memory as Pandas DataFrames or NumPy arrays. This design is incompatible with deep learning workflows, where data must be processed in small batches, forcing practitioners to abandon popular toolkits and manually reimplement fairness methods.

The second issue is methodological: current tools primarily offer quantitative metrics or basic visualization capabilities, which are insufficient for analyzing biases in computer vision models. Furthermore, the post-hoc debiasing techniques they provide typically operate only on the model's outputs - such as by adjusting classification thresholds - without addressing or removing bias in the model’s underlying reasoning process.

To address these gaps, we introduce DetoxAI, a post-hoc debiasing toolkit for image classification that operates at the representation level. DetoxAI enables desensitization of neural networks to protected attributes (e.g., gender, race) without requiring full retraining. Designed for deep learning and seamlessly integrated with PyTorch, DetoxAI equips AI practitioners and researchers with a practical tool for empirical evaluation, comparative studies, and the deployment of fairness interventions in real-world vision models.
\begin{figure}[tp] 
\centering 
\includegraphics[width=0.85\linewidth, trim=0.5cm 1.40cm 1.6cm 0.6cm, clip]{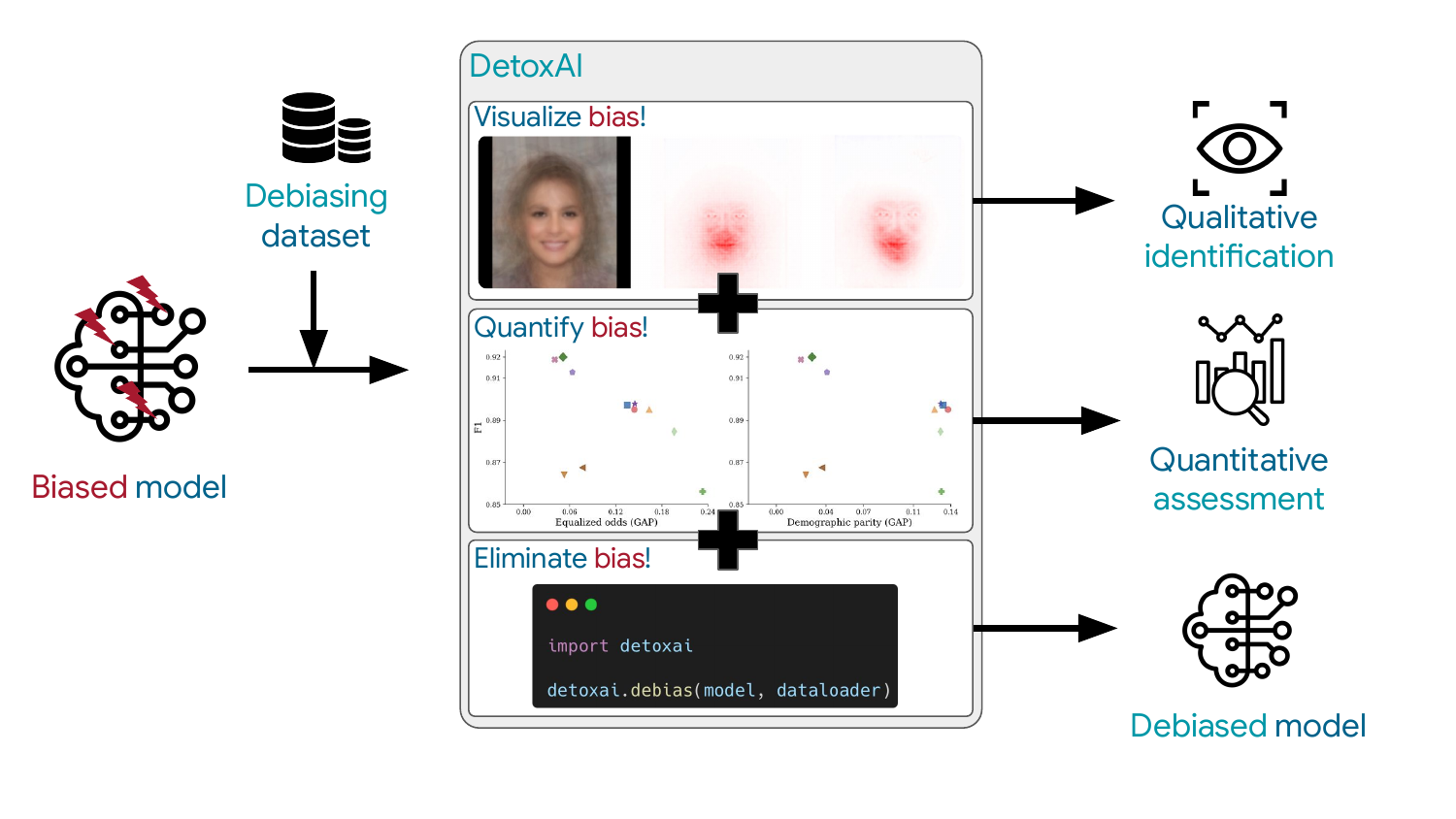} \caption{DetoxAI incorporates multiple tools to mitigate biases in vision models. Biased model with debiasing dataset is passed to our module, where biases can be inspected, measured and finally eliminated.} 
\label{fig:workflow} 
\end{figure}

% \section{Use Cases: DetoxAI in Engineering and Research}
\section{Use Cases}

\paragraph{\bfseries Engineering Use Case: Debiasing a facial expression recognition system}

Consider a facial expression recognition system deployed to detect whether customers are smiling. In practice, it is observed that the system consistently misclassifies individuals wearing neckties as not smiling, revealing unintended bias. Using DetoxAI, engineers can address this by treating the necktie as a protected attribute. DetoxAI allows them to load the existing model (e.g., an already trained and deployed ResNet50) and apply targeted debiasing techniques - all without full model retraining. The updated model can then be seamlessly redeployed, mitigating the bias while maintaining overall performance. Engineers can evaluate fairness metrics, visualize attribution shifts, and select the optimized model, enhancing both fairness and system reliability.

\paragraph{\bfseries  Research Use Case: Fairness studies and comparative benchmarking}

Researchers studying fairness can leverage DetoxAI as a standardized platform for systematic benchmarking. DetoxAI enables easy comparison of post-hoc debiasing methods under consistent pipelines, evaluation on common fairness metrics, and qualitative analysis of shifts in feature importance. Its modular design and clear abstractions simplify the integration of novel debiasing methods and benchmarking against established techniques. This flexibility can help accelerate fairness research in image classification and supports rigorous, extensible and reproducible experimentation.

\begin{figure}[tp] 
\centering 
\hspace{-1cm}
\includegraphics[width=0.9\linewidth]{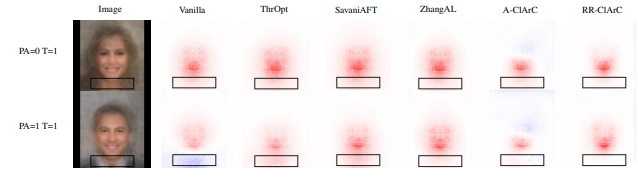}
% \vspace{1cm}
\includegraphics[width=0.85\linewidth]{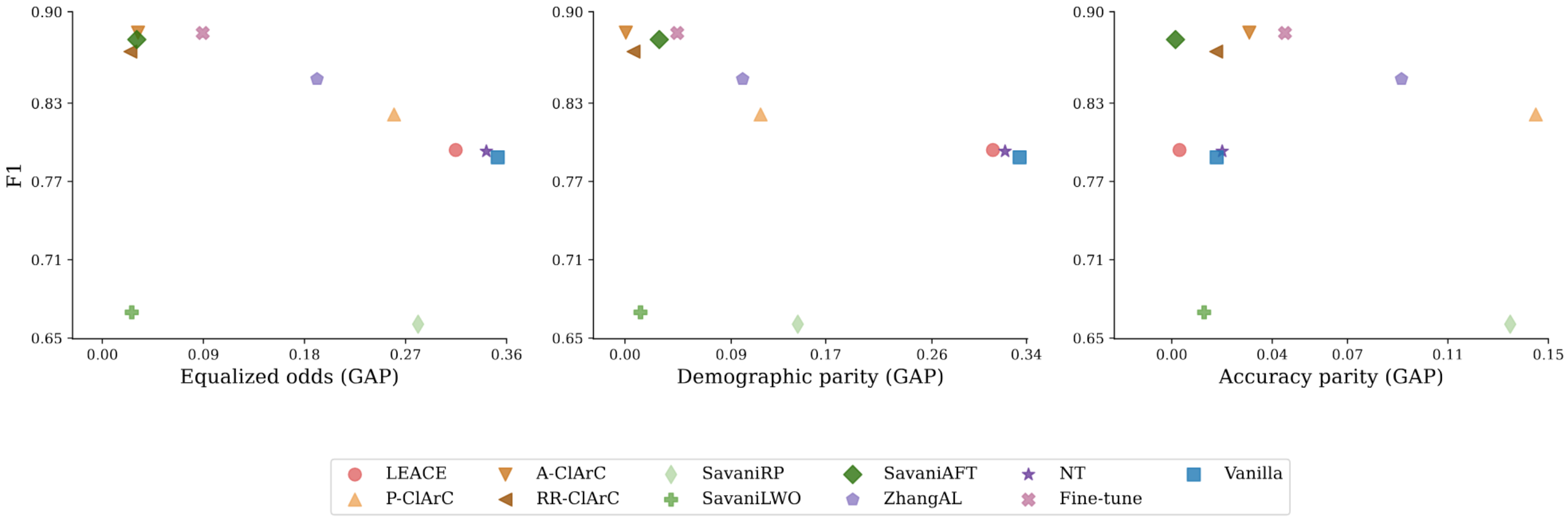} 
\caption{In the upper image, we visualize saliency maps before (Vanilla) and after bias mitigation using a selection of implemented methods. The image on the bottom is a quantitative evaluation of the debiasing techniques on the predictive performance (F1 Score) - fairness trade-off. Fairness metrics are EqualizedOdds, DemographicParity and AccuracyParity, calculated in a difference between protected groups variation \cite{mehrabi_2021_survey}.} \label{fig:relevance} 
\end{figure}

\section{System Overview}

DetoxAI implements several post-hoc debiasing methods, including Savani and Zhang based methods \cite{savani}, LEACE \cite{leace}, and post-hoc Threshold Optimization \cite{mehrabi_2021_survey} It also includes ClArC variants \cite{clarc} originally used as artifact-removal methods but repurposed for fairness \cite{sztukiewicz2025investigating}. All methods, except Threshold Optimization, modify internal representations instead of merely calibrating outputs, enabling deeper mitigation of learned biases.

Debiasing techniques are accessed through a unified \texttt{detoxai.debias(...)} interface (Fig.~\ref{fig:workflow}). The API supports debiasing, performance evaluation (e.g., F1, GMean, Balanced Accuracy), and fairness metric computation (e.g., Equalized Odds, Demographic Parity, Accuracy Parity), offering seamless integration into existing PyTorch workflows.

DetoxAI is model-agnostic and focuses on binary classification tasks with binary protected attributes. Even though certain debiasing methods require internal model interventions (e.g., hooks) and fine-tuning, DetoxAI automatically adapts to a variety of PyTorch models without requiring additional user input. All debiasing methods come with default configurations, empirically tuned for robustness across various model sizes and architectures. Advanced users can override these defaults by passing custom configurations to the API to better meet specific needs, such as computational budget or optimized fairness metrics. All components follow an object-oriented design, making it straightforward to extend DetoxAI with new methods, metrics, or visualization tools (see Sec.~\ref{sec:links} for examples and documentation).

\section{Conclusions}
DetoxAI offers a unified platform for implementing fairness interventions in deep learning systems for image classification tasks. Designed to be production-ready, yet highly extensible, the toolkit supports a wide range of practical applications across both industry and research. Its simple and consistent API lowers the barrier to applying bias mitigation techniques, making fairness interventions accessible to engineers and researchers alike. 
%DetoxAI aims to promote broader consideration of fairness as a standard component of machine learning workflows.

\section*{Additional Information}
\label{sec:links}
Webpage and documentation: \url{https://detoxai.github.io},    
GitHub: \url{https://github.com/DetoxAI/detoxai}.

\section*{Acknowledgments} 
The work was supported by a Ministry of Science and Higher Education grant (No. MNiSW/2025/DPI/56) as part of the FERS programme, co-financed by the European Union. Jerzy Stefanowski was supported by a National Science Centre grant (No. 2022/47/D/ST6/01770).

\bibliography{ref}
\bibliographystyle{splncs04}

\end{document}